\documentclass[11pt,a4paper]{article}
\usepackage[hyperref]{acl2020}
\usepackage{times}
\usepackage{latexsym}

\usepackage{microtype}

\aclfinalcopy % Uncomment this line for the final submission
\usepackage{graphicx}
\usepackage{amsmath}
\usepackage{amssymb}
\usepackage{amsthm}
\theoremstyle{definition}
\newtheorem{definition}{Definition}[section]
\usepackage[ruled,linesnumbered]{algorithm2e}
\usepackage{algorithmic}
\usepackage{multirow}
\usepackage{subcaption}
\usepackage[normalem]{ulem}
\usepackage{arydshln}  % for \hdashline
\usepackage{hyperref}
\newcommand{\tn}[1]{\textnormal{#1}}

\title{Detecting and Understanding Generalization\\ Barriers for Neural Machine Translation}

\author{Guanlin Li\textsuperscript{$\epsilon$}~\thanks{~~~Preprint; work done at Tencent AI Lab.}~~, Lemao Liu\textsuperscript{$\lambda$}, Conghui Zhu\textsuperscript{$\epsilon$}
, Tiejun Zhao\textsuperscript{$\epsilon$}, Shuming Shi\textsuperscript{$\lambda$} \\
\textsuperscript{$\epsilon$}Harbin Institute of Technology, \textsuperscript{$\lambda$} Tencent AI Lab \\
    \{epsilonlee.green\}@gmail.com, \{chzhu, tjzhao\}@hit.edu.cn, \\
    \{redmondliu, shumingshi\}@tencent.com \\
}

\date{}

\begin{document}
\maketitle
\begin{abstract}

Generalization to unseen instances is our
eternal pursuit for all data-driven models.
However, for realistic task like machine translation,
the traditional approach measuring generalization
in an average sense provides
poor understanding for the fine-grained generalization ability.
% especially under trustworthy consideration.
As a remedy,
this paper attempts to identify and understand generalization
barrier words within an unseen input sentence
that \textit{cause} the degradation of fine-grained generalization.
We propose a principled definition of generalization barrier words and a
modified version which is tractable in computation.
Based on the modified one, we propose three simple methods
for barrier detection by the search-aware risk estimation
through counterfactual generation.
We then conduct extensive analyses on those detected generalization
barrier words on both Zh$\Leftrightarrow$En
NIST benchmarks from various perspectives.
Potential usage of the detected barrier words
is also discussed.

\end{abstract}

\section{Introduction}
The performance of neural machine translation (NMT)
models have been boosted significantly through novel architectural attempts~\citep{gehring2017convolutional,vaswani2017attention},
carefully-designed learning strategies~\citep{Ott2018ScalingNM}
and
% is significantly boosted via
semi-supervised techniques
that smartly increase the size of training
corpous~\citep{Edunov2018UnderstandingBA,ng2019facebook}.
Meanwhile, by leaving architectural choice unchanged,
empirical result shows that simply increasing the
model capacity via delicate gradient control can lead
to faster convergence and better performance~\citep{wang2019learning,biaozhang-etal-2019-improving}.
% However, these two empirical principles, i.e. increasing data size,
% increasing model size, can hinder scientific understanding of the
% capability of a model trained on a given dataset.
However, all these improvements are measured in an \textit{average}
sense on a held-out dataset and two potential limitations may stand out.
On the one hand, we should be careful
about this average test performance comparison paradigm
due to issues like
test set overfitting~\citep{pmlr-v97-recht19a,NIPS2019_9190};
on the other hand, the average
case analysis only covers the mean data population
and does not provide much information on
questions like what properties
of the unseen input hinders model's generalization, which are receiving
great attention in the trustworthy machine learning community~\citep{Amodei2016ConcretePI,jia-etal-2019-certified}.

One possible solution to mitigate the above limitations
is to analyze the property of
the unseen input sentence as a whole,
which is an instance-level analysis
instead of average analysis. This is similar to recent
renaissance of out-of-distribution detection in the task of
image classification~\citep{chandola2009anomaly,hendrycks17baseline,Liang2017EnhancingTR}.
However, since for the task of machine translation an input sentence consists of many words, we find that
the overall generalization
of the model on the sentence is mostly effected by a few words
and modifying them can improve translation quality largely.
This phenomenon is shown in Figure~\ref{fig:example-risk-dist}, where
by changing \textit{qu\=exi\`an} to \textit{g\=enj\`u}, the input sentence is translated much better instead.
Therefore, it would be more appropriate to analyze
those generalization barrier words, e.g. the words within an input sentence which hinder
the overall generalization of the model on that sentence.
%We call those substructures as generalization barriers.

To this end,
we firstly give a principled definition of generalization barrier in a
counterfactual~\citep{Pearl:2018:BWN:3238230} way. Since the principled definition requires human evaluation, we instead provide a modified definition based on a novel statistics, which employs automatic evaluation to detect
generalization barrier words.
%The approximate definition of generalization barrier is measured by a novel statistics from counterfactuals.
%\lml{We use statistics from counterfactuals to annotate each source word with a risk of it being the generalization barrier.}
As it is costly to exactly compute this statistics, we propose three approximate estimators to inexactly calculate its value.
In terms of the calculated value, we conduct experiments on two benchmarks to detect potential barriers in each unseen input sentence. In addition, we carry out systematic analyses on the detected barriers from
different perspectives. We find that generalization barrier words are pervasive
among different linguistic categories (Part-of-Speech) and very different from
previously known troublesome source words~\citep{zhao2018addressing,zhao2019addressing}.
Generalization barrier words tend to be complementary
across different architectural choice. Moreover, modification of
barrier words leads to more diversified
hypo candidates which might be a better
choice for re-ranking~\citep{yee-etal-2019-simple}
than the top-$k$ outputs under one steady input via beam search.

\section{Related Literature}

\noindent
\textbf{Troublesome words detection}
To our knowledge, back to the old SMT era, \citet{mohit-hwa-2007-localization} is the
most related work which invents the notion of
'hard-to-translate phrase' at source side,
and uses removal to determine its effect on model generalization on
other phrases' translation,
which is very similar to our usage of
counterfactual generation by editing the source words.
Recently, \citet{zhao2018addressing,zhao2019addressing} are the first
to detect \textit{trouble makers} at source side globally for NMT. In
\citet{zhao2018addressing}, the troublesome source words are detected through
an exception rate defined as the number of troublesome alignments $(x_i, y_j)$
dividing the number of $x_i$, where the troublesome alignments are obtained
through an extrinsic statistical aligner instead of the trained NMT model.
In \citet{zhao2019addressing}, the troublesome source words are constrained to
words with high translation entropy which tend to be under-translated by the
model. Both of their trouble detection heuristics are: 1) context-unware,
globally applied on every source words without considering the context of the
words, and 2) model-unaware, dependent on extrinsic statistical assumptions.
In our work, we are trying to detect both
context-aware and model-specific generalization
barriers for every unseen source input.

\noindent
\textbf{Out-of-Distribution (OOD) detection}
OOD detection, Novelty~\citep{markou2003novelty},
Outlier~\citep{hodge2004survey} or Anomaly Detection~\citep{chandola2009anomaly}
care about how likely the unseen input \textit{as a whole}
is to be sample different from the
training distribution. This problem is recently revived on the task of image
classification~\citep{hendrycks17baseline,Liang2017EnhancingTR,Choi2018WAICBW}.
Although recently, \citet{Ren2019LikelihoodRF} starts to consider OOD
detection on sequential data, i.e. gene fragments, they still regard the input
feature as a holistic vehicle to cause the mismatch in underlying generative
distribution. Our work is motivated from this OOD detection literature in the
spirit of detecting the inputs that the model cannot generalize well upon. Beyond that,
due to the structural property of the translation task, we also carry out a more
fine-grained detection of \textit{causes}
that could be a part of the input feature,
which can potentially consist of several high risky words. Notably, researchers
from OOD detection recently start to focus on structure of the
input and design benchmarks for such detection task for image
anomaly segmentation which focuses on
small patches in the image~\citep{hendrycks2019anomalyseg}.

% \noindent
% \textbf{Uncertainty modeling}
% Uncertainty is a surrogate of model's performance, which can be used to reflect
% the capabilities of the model. In OOD detection, uncertainty or confidence is
% used as indicator of novelty~\citep{hendrycks17baseline,Liang2017EnhancingTR}.
% In \citet{dong-etal-2018-confidence}, they are the first to systematically adopt
% the uncertainty view from \citet{Gal2015DropoutAA,Kendall2017WhatUD} and connect
% several measures of uncertainty directly to the performance of a neural semantic
% parser. They also propose to propogate the model uncertainty back to the input
% tokens to attribute the source of uncertainty. This motivates us to consider a
% direct relationship between the input and the predictive performance of an NMT
% model. Morevover, we also provide in-depth understanding of the detected
% generalization barriers. In our reference-free methods, we use uncertainty
% to accomplish hypothesis risk annotation, which is lightweight and simple.

\noindent
\textbf{Error analysis and interpretability}
Recently, \citet{wu-etal-2019-errudite} propose to conduct error analysis with
three principles by heart: scalable, reproducible and counterfactual
for natural language processing tasks.
These principles also guide the computational
consideration of our detection method. For NMT, recently,
% as far as we know,
\citet{lei-etal-2019-revisit} are the first to focus on accurately detecting
wrong and missing translation of certain source words.
Different from their work which
detects the unsatisfactorily translated source words themselves,
our work focuses on detecting the \textit{cause} of them, and serves
as complementary to recent interpretability analysis of importance
words~\citep{he-etal-2019-towards}.

\section{Generalization Barriers}

\begin{figure*}[t]
    \centering
    \includegraphics[scale=0.57]{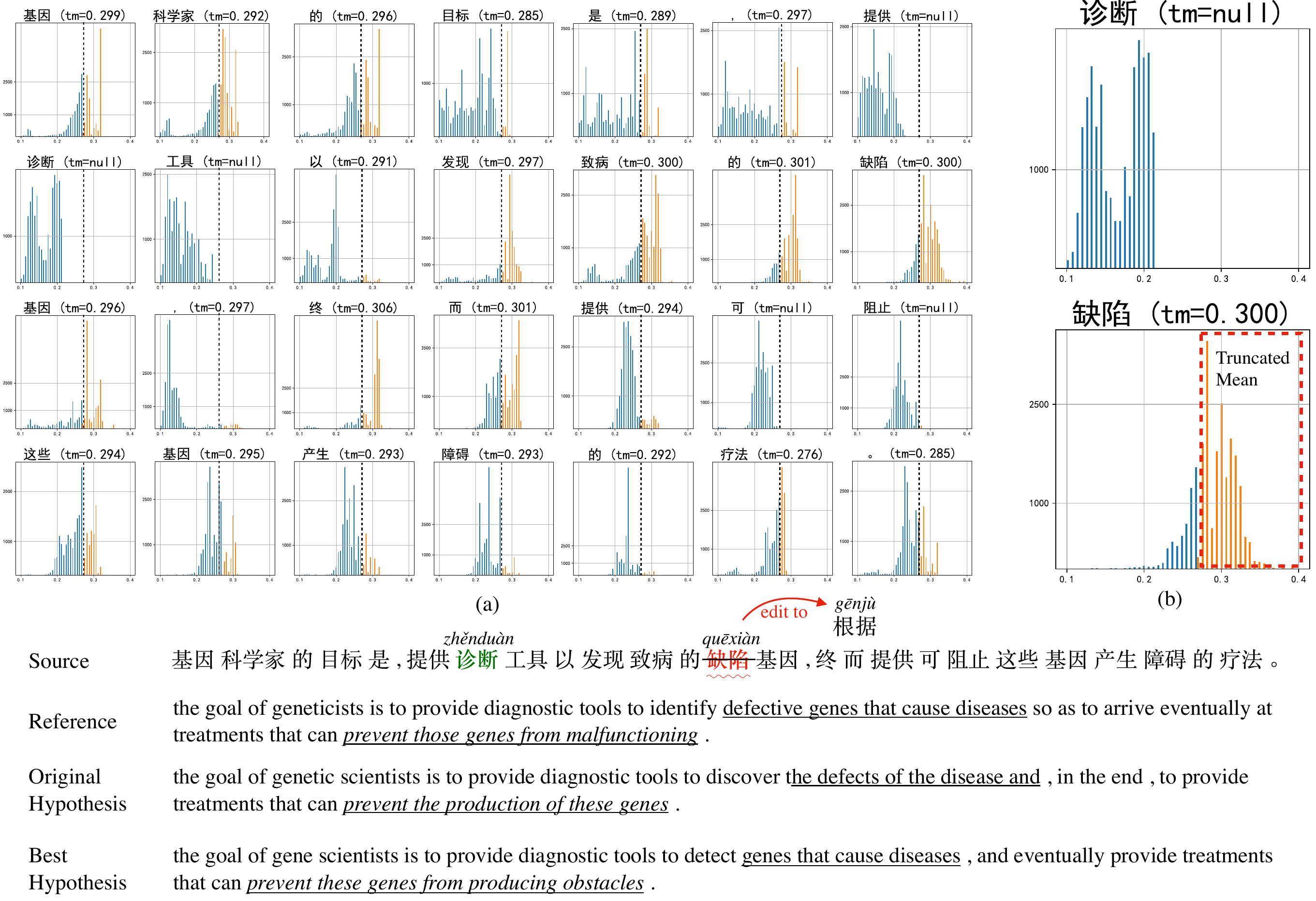}
    \caption{A showcase of histograms of the evaluation metric value (smoothed sentence-level BLEU) at all source positions. Every histogram is drawn by collecting every possible metric value after editing ($\vert \mathcal{V} \vert$ values in total), and then BLEU spectrum 0.1-0.4 on the x-axis is divided into 50 bins. In each histogram, the orange band (if exists) shows the metric values of the edited sources that are above than the original metric value, the blue part shows the metric values that are below the original metric value. We can judge the risk of a word being a generalization barrier word by focusing on the orange band, i.e. we measure the its truncated mean as an average risk of the word.
    % The visualizations of those performance distributions at different source positions motivate us to use the truncated mean (the mean of the orange band) as the risk of the positions of being generalization barrier words.
    }
    \label{fig:example-risk-dist}
\end{figure*}

Mainstream NMT is formulated as a sequence-to-sequence structured prediction
problem and modeled and factorized as follows: % with directional fully-connected target-side label dependency:
\begin{equation}
    P(\tn{y} \vert \tn{x}; \theta) = \Pi_j P(\tn{y}_j \vert \tn{x}, \tn{y}_{<j}; \theta),
\end{equation}
Maximum Likelihood Estimation (MLE) training is conducted on the training set
$\mathcal{D}^{tr} = \{ (\tn{x}^i, \tn{y}^i) \}$,
where $\tn{x} \in \mathcal{X}$, $\tn{y} \in \mathcal{Y}$, with minibatch SGD
to obtain an estimation of the parameter weights $\hat{\theta}$
\citep{Luong2015StanfordNM,gehring2017convolutional,vaswani2017attention}.
Like all other structured prediction problems with a scoring function
and a decoding algorithm~\citep{daume06thesis},
for NMT, $P(\tn{y} \vert \tn{x}; \theta)$ acts as the scoring
function and beam search is used as the (approximate) decoding algorithm.
Since beam search is a deterministic algorithm
with a preset beam size, the prediction $\hat{\tn{y}}$ is
\textit{solely} determined by the input $\tn{x}$, denoted as a map
$\hat{\tn{y}} = \mathcal{M}_{\hat{\theta}}(\tn{x})$.
Under this setting,
we are interested in the causal question: \textit{how the input $\tn{x}$
causes the model's failure on the prediction?}

As natural language sentences, the input
$\tn{x} = (\tn{x}_1, \dots, \tn{x}_m)$ is a sequence
with compositional structure that forms the whole semantics of itself.
% Since compositionality is a complex concept~\citep{sep-compositionality},
% here we do not constrain the specific compositional structure
% of $\tn{x}$ but use constituency parse as a showcase in Figure~\ref{xxx}.
On the one hand, we want the NMT model to
generalize well on the previous unseen input $\tn{x}$, which means
ideally it should be able to generalize well on any possible (meaningful)
subsequences of $\tn{x}$.
On the other hand, the cause of the model's generalization
degradation should be attributed to some of the
subsequences or their ways of composition.
Therefore, one perspective to
shed light on the above \textit{how} question
is to try to \textit{detect} the set of
all subsequences of $\tn{x}$ that
can potentially deteriorate
model's generalization, which we dub \textit{generalization barriers}.
We give a principled but
abstract definition of generalization barriers and its
approximate but tractable version
in the following subsections by treating each
source word independently without considering their
possible combinatorial compositions.
Then we construct a statistics for each source word
to represent its risk of being a generalization barrier word.

% \subsection{The principled definition}
\subsection{A definition with human effort}

The principled definition of generalization barriers
is based on the intuition that the model can potentially
generalize well on some edited versions of $\tn{x}$,
i.e. with \textit{words substitution} and \textit{deletion} that
try to partially preserve the original
symbolic compositional structure (e.g. word order)
and semantics of $\tn{x}$ as much as possible.
This intuition also
matches with the causal question we have asked before,
since we are actually generating
counterfactuals through intervening (editing) $\tn{x}$~\citep{Chang2018ExplainingIC}.
% Specifically, if we denote a certain edited version of $\tn{x}$ as
% $\tilde{\tn{x}}^k$, the generalization barrier is
% defined as $(\tn{x} \setminus \tilde{\tn{x}}^k)$,
% where $k$ indexes different edited versions.
% Here, the operator $\setminus$ between sequence $\tn{x}$
% and $\tilde{\tn{x}}^k$ returns a subsequence of $\tn{x}$,
% denoted as
% $\tn{x}^{sub} = (\tn{x}_{l_1}, \tn{x}_{l_2}, \dots, \tn{x}_{l_{m'}})$,
% where we have $1 \leq l_1 < l_2 < \dots < l_{m'} \leq m$.
Specifically, if we denote a certain edited version of
$\tn{x}$ as $\tilde{\tn{x}}$, the generalization barrier is
defined as $(\tn{x} \setminus \tilde{\tn{x}})$ , where $\tilde{\tn{x}}$ can be any edited version of $\tn{x}$ if $\tilde{\tn{x}}$ satisfies the constraints
in Definition~\ref{def:generalization-barriers}.
The operator $\setminus$ returns a subsequence of
$\tn{x}$ by removing their overlapped words.

% $\tn{x}^{sub} = (\tn{x}^{sub}_1, \tn{x}^{sub}_2, \dots, \tn{x}^{sub}_l)$,
% where $\tn{x}^{sub}_j \in \tn{x}, \tn{x}^{sub}_j \notin \tilde{\tn{x}}^i$.
% We denote one possible reference translation of $\tilde{\tn{x}}^i$
% as $\tilde{\tn{y}}^i$, which can be provided by human annotators.
% We use a metric $m$ to measure the translation quality
% of $\mathcal{M}_{\hat{\theta}}(\tilde{\tn{x}}^i)$.

\begin{definition}{(\textit{Generalization Barriers})}
\textit{
Given an NMT model trained on $\mathcal{D}^{tr}$ with $\hat{\theta}$,
a distance measure
$d: \mathcal{X} \times \mathcal{X} \mapsto \mathbb{R}$,
e.g. the edit distance,
for an input $\tn{x}$, we call the set of subsequences,
$\cup (\tn{x} \setminus \tilde{\tn{x}})$,
that satisfy the following constraints as generalization barriers of $\tn{x}$.
% For each subsequence $(\tn{x} \setminus \tilde{\tn{x}}^k)$, it should
% satisfies:
% solution of the following optimization problem as $\tilde{\tn{x}}$, the generalization barriers of the given input as $\tn{x} \setminus \tilde{\tn{x}}$, where $m(\cdot, \cdot)$ could be any evaluation metric or the human oracle.
\begin{enumerate}
    \item The distance measure $d(\tn{x}, \tilde{\tn{x}})$ is minimized;
    \item {
    % $m(\tilde{\tn{y}}^i, \mathcal{M}_{\hat{\theta}}(\tilde{\tn{x}}^i)) > \tau$
    % , or
    Human evaluation of the translation quality on
    $\mathcal{M}_{\hat{\theta}}(\tilde{\tn{x}})$ reaches a satisfactory level.}
\end{enumerate}
}
% \begin{align}
% & \text{minimize}_{\tn{x}' \in \mathcal{X}}           &  \text{EditDistance}(\tn{x}', \tn{x}) \nonumber \\
% & \text{subject to}         &  m(\mathcal{M}_{\hat{\theta}}[\tn{x}'], \tn{y}') > \tau \nonumber \\
% \label{eq:gb-minimize}
% \end{align}
\label{def:generalization-barriers}
\end{definition}

\noindent
\textbf{Remarks} Definition~\ref{def:generalization-barriers} is principled because:
a) it respects the compositional nature of possible generalization
barriers instead of considering only individual word;
b) it handles semantic shift properly by largely preserving
the original words and the word orders through minimized $d$.
The second benefit can be seen as false discovery
control~\citep{pmlr-v97-gimenez19a},
since without this distance constraint, we can always find
a well-generalizable subsequence in $\tn{x}$ by deleting
most of the words.
% though some of them can be reliably
% translated with the remained words as well.
% to control the false positive rate
% of detection while guarantees the recall of generalization barriers.

% \subsection{The modified definition}
\subsection{Approximating the definition with counterfactuals}

% However, the above definition is also hard to scale up when
% NMT system developers want to conduct large-scale automatic
% error analyses~\citep{wu-etal-2019-errudite}
% due to the overuse of human evaluation.
However, the above definition is also hard to scale up
due to large search space and human evaluation. So we
further make the following assumptions to modify
it: a) the minimization of $d$ is purposefully set
to $d=1$, which restricts the search space tremendously
by only editing one word for investigating its possibility
of being a barrier;
b) the human evaluation is replaced
by automatic evaluation with a metric such as
smoothed sentence-level BLEU~\citep{Lin2004AutomaticEO},
since $d=1$ roughly leads to
an unchanged reference $\tn{y}$.

Now we investigate each source word $\tn{x}_i$ independently
by counterfactual generation as well. Instead of
finding one single counterfactual $\tilde{\tn{x}}$
which might be unsuitable
for human to perceive as a natural sentence, inspired
by \citet{burns2019interpreting} and \citet{Chang2018ExplainingIC}
who edit certain patch in an image with potentially
infinitely infilling patches and compute importance
score of the original patch
in expectation, we also generate as many
edit choices as possible so that many edits may include a natural sentence. Suppose $\mathcal{V}$ is the source vocabulary, $\text{Edit}(\tn{x}, i)$ is the set of all sentences edited from $\tn{x}$ at position $i$.
Accordingly, the size of $\text{Edit}(\tn{x}, i)$ is $\vert \mathcal{V} \vert$, which corresponds to one deletion and $\vert \mathcal{V} \vert$ substitutions.
Then we can actually obtain
$\vert \mathcal{V} \vert$ counterfactual performance measures:
\begin{equation}
    \mathcal{S} = \{ \text{BLEU}(\mathcal{M}_{\hat{\theta}}(\tilde{\tn{x}}), \tn{y}) \vert \tilde{\tn{x}} \in \text{Edit}(\tn{x}, i) \},
    \label{eq:s}
\end{equation}
\noindent
% As a solution, to take advantage of the original reference translation $\tn{y}$,
% we restrict the distance $d$ to be small enough so that
% the edited version of $\tn{x}$ can be treated as having the same translation
% reference. Specifically, we \textit{only edit one word per time}
% in $\tn{x}$ to \textit{independently} determine
% whether that word is harmful to the model's generalization on $\tn{x}$.
% Since we can delete $\tn{x}_i$ or substitute it with any other word in the
% source vocabulary,
% in total, there are $\vert \mathcal{V}^s \vert$ possible editing choices
% which lead to $\vert \mathcal{V}^s \vert$
% potential counterfactual outcomes
% $\mathcal{M}_{\hat{\theta}}(\tilde{\tn{x}}^k)$.
% Under certain automatic evaluation metric
% $m: \mathcal{Y} \times \mathcal{Y} \mapsto \mathbb{R}$,
% e.g. smoothed sentence-level BLEU~\citep{Lin2004AutomaticEO},
% we can thus collect
% a set of $\vert \mathcal{V}^s \vert$ metric values
% given the reference $\tn{y}$ as:
% \begin{equation}
%     \mathcal{S} = \{ m^k \vert m^k = m(\mathcal{M}_{\hat{\theta}}(\tilde{\tn{x}}^k), \tn{y}) \},
% \end{equation}
% We draw a \textit{histogram}
% with different bins of metric value intervals.
based on which we can draw a \textit{histogram}
with binned metric values.
Figure~\ref{fig:example-risk-dist} is a showcase
for a given input sentence,
we conduct $\vert \mathcal{V} \vert$ real decoding
for each of the 28 words and plot the corresponding histograms
which have tremendous information.
We regard each histogram as a distribution
of the counterfactual
generalization performances.

As we can identify in Figure~\ref{fig:example-risk-dist},
the right orange band of a histogram
(if exists) shows the counterfactuals
with better generalization, and if that part \textit{dominates}
the distribution, we can conclude that the word being
edited has a \textit{high risk} of \textit{causing}
the degradation of generalization on $\tn{x}$.
In practice, we use the empirical truncated mean
at position $i$
to represent $\tn{x}_i$'s risk of being a generalization
barrier word as follows:
\begin{equation}
    tm(\tn{x}_i) = \frac{1}{\vert \mathcal{S}_{> m^o} \vert} \sum_{v \in \mathcal{S}_{> m^o}} v,
\label{eq:truncated-mean}
\end{equation}
where
$\mathcal{S}_{> m^o} = \{ v \vert v > m^o, v \in \mathcal{S} \}$ and
$m^o = \text{BLEU}(\mathcal{M}_{\hat{\theta}}(\tn{x}), \tn{y})$.
The set $\mathcal{S}_{> m^o}$ corresponds
to the orange band in the histogram.
The higher the risk, the more likely that word being a
generalization barrier word.
In Figure~\ref{fig:example-risk-dist},
the truncated mean (tm) is shown above each histogram, with
'null' denotes that position has no orange band.
% Based on the observation,
% we give a practical definition of generalization
% barrier words.
\begin{definition}{(\textit{Generalization Barrier Words})}
\textit{
The generalization barrier words in $\tn{x}$ are those
$\tn{x}_i$ whose $tm(\tn{x}_i)$ reaches a satisfactory level $\tau$.
% are the words with top-$k$ highest truncated means through Eq.~\eqref{eq:truncated-mean} while being edited independently.
}
\label{def:generalization-barrier-words}
\end{definition}
\noindent
In practice the hard threshold $\tau$
could vary for different $\tn{x}$,
so we use a soft one, the top-$k$ risky words,
for deciding the potential generalization barriers.

\noindent
% Although definition~\ref{def:generalization-barrier-words} is more tractable and does not
% involve human effort, it is also computationally inefficient due to the $\max$
% operator over all possible substitutions along the source vocabulary,
% where we need to call the decoding algorithm $\vert \mathcal{V}^{src} \vert$ times,
% usually above ten thousands large.

% \begin{figure}[t]
%     \centering
%     \includegraphics[scale=0.65]{fig/overlapk.pdf}
%     \caption{The Overlap@$k$ of different budgets on different samples; for each budget, the concentrated mean can be seen as the average overlap rate of that budget.\color{red}{TODO: use boxplot to replace scatter plot.}}
%     \label{fig:overlap-at-k}
% \end{figure}

\begin{algorithm}[htb]
\begin{algorithmic}[1]  % 1 means show psudocode line number
\small{
\REQUIRE ~~~ \\  % Input
A risk estimator $\text{S}$;\\% the metric $m(\cdot, \cdot)$;\\
an unseen pair $\tn{x}, \tn{y}$, position $i$, budget $B$, $b$;\\
the learned NMT model $P(\tn{y} \vert \tn{x}; \hat{\theta})$,\\
the source embedding
$\text{Emb} \in \mathbb{R}^{\vert\mathcal{V}\vert \times d}$;\\
%the original metric value $m^o$,
% budget $B$, $b$;\\
\ENSURE ~~~ \\ % Output
The estimated truncated mean $tm(\tn{x}_i)$;
% start of the body
\STATE Initialize $\text{C}^i = \{\}$;
%\STATE Add $\tn{x}_{\setminus i}$ to $\text{C}^i$; $//$ \textit{for deletion}\\
\IF{S = \textit{Uniform}}
% \STATE $//$ \textit{uniform sampling}
\STATE Uniformly sample $b$ elements from $\textrm{Edit}(\tn{x}, i)$,\\and add them to $\text{C}^i$;
\ELSIF{S = \textit{Stratified}}
% \STATE $//$ \textit{stratified sampling}
\STATE Uniformly sample $B$ elements from $\textrm{Edit}(\tn{x}, i)$ as $\text{C}^i_0$;
\STATE Compute $\mathcal{L}_{\hat{\theta}}(\tilde{\tn{x}})$ in Eq.\eqref{eq:str-loss} for each $\tilde{\tn{x}} \in \text{C}^i_0$;
\STATE Use $s_{\tilde{\tn{x}}} \propto 1/\mathcal{L}_{\hat{\theta}}(\tilde{\tn{x}})$ to choose the top-$b$ elements\\in $\text{C}^i_0$, and add them to $\text{C}^i$;
\ELSIF{S = \textit{Gradient-aware}}
% \STATE $//$ \textit{gradient-aware sampling}
\STATE Compute Eq.~\eqref{eq:one-step-grad} to get $\text{Emb}'(\tn{x}_i)$;
\STATE Use $\text{softmax}(\text{Emb} \cdot \text{Emb}'(\tn{x}_i))$ to sample $b$\\elements from $\textrm{Edit}(\tn{x}, i)$, and add
them to $\text{C}^i$;
\ENDIF
% end of the body
% \STATE $//$ \textit{real decoding to compute} $\max$\
\STATE Conduct real decoding on $\text{C}^i$ and compute\\$tm(\tn{x}_i)$ supported on $\text{C}^i$ rather than $\textrm{Edit}(\tn{x}, i)$.
\RETURN $tm(\tn{x}_i)$;
}
\end{algorithmic}
\caption{Evaluate the risk of $\tn{x}_i$}
\label{alg:eval-risk}
\end{algorithm}

\subsection{Estimating the truncated mean}

% Although Definition~\ref{def:generalization-barrier-words} no longer
% counts on human effort, the time cost of $\vert \mathcal{V}^s \vert$
% times real decoding for every $\tn{x}_i$ still hinders
% large-scale analysis. Therefore, we would like to limit the number
% of real decoding under a budget $b$
% which is far less than $\vert \mathcal{V}^s \vert$.
% Note that since we always select deletion as one
% of $b$, there are $b - 1$ choices for substitutions.

% Hypothetically, if we regard the metric value at each $\tn{x}_i$
% as a random variable $M$, and
% each histogram in Figure~\ref{fig:example-risk-dist}
% as governed by a density function $f(m)$ that maps metric value to its density
% (frequency), and the cumulative distribution
% function is denoted as
% $F(m)$.~\footnote{{\url{https://en.wikipedia.org/wiki/Truncated_distribution}}}
% Given the metric value on the original hypothesis $m^o$, the truncated density
% function is denoted as $f^{tr}(m \vert M > m^o) = \frac{f(m)}{1 - F(m^o)}$,
% based on which we want to estimate
% its mean which is the truncated mean of $f(m)$.
% $f^{tr}$ is unknown but can be estimated through
% sampling from $f$ to depict an approximate histogram.
% From this stochastic view, we can think of
% \textit{uniformly} sampling from the $\vert \mathcal{V}^s \vert$
% edit choices as sampling from $f$.

%Generally, it is still intractable to exactly calculate $tm(\tn{x}_i)$.

According to the definition in Eq.\eqref{eq:truncated-mean} and Eq.\eqref{eq:s}, one has to decode each $\tilde{\tn{x}}\in \textrm{Edit}(\tn{x},i)$ and there are $|\mathcal{V}|$ sentences in total.  Unfortunately, as it takes a few seconds for each decoding, it is impractical to exactly calculate $\mathcal{S}$ as well as $\mathcal{S}_{>m^o}$
 %$\mathcal{S}_{>m^o}$ for exactly calculating $tm(\tn{x}_i)$. Unfortunately, since the distribution of $\hat{x}$ is unknown and there are $$ $\hat{x}$, the cardinality of $S$ is exponential and thus that of $\mathcal{S}_{>m^o}$ may be exponential as well. Therefore, it is intractable to calculate the exact $tm(\tn{x}_i)$.
As a result, we instead propose a simple yet effective algorithm as an inexact solution.
The key idea to the inexact solution is to call the decoder $b$ times, with $b$ as a budget.
Specifically, we randomly sample $b$ elements from $\textrm{Edit}(\tn{x},i)$ to obtain a sample set $\text{C}^i$. Then we calculate both $\mathcal{S}$ and $\mathcal{S}_{>m^o}$ supported on $\text{C}^i$. Finally we can approximately calculate $tm(\tn{x}_i)$ by enumerating at most $b$ elements in $\mathcal{S}_{>m^o}$. To randomly sample $b$ elements from $\textrm{Edit}(\tn{x},i)$, we predefine three distributions heuristically, which lead to three different estimators as follows.

% Therefore, we assume a budget of $b$ times
% of real decoding which controls the computation burden.
% Here, we propose to use a very simple
% unbiased estimator of the truncated mean
% through Monte Carlo sampling: we uniformly sample from vocabulary with
% certain budget $b$, and compute the truncated mean
% as Eq.~\eqref{eq:truncated-mean} but with fewer samples.
% We conduct experiments on 5 inputs with 159 words in total to show
% the trade-off of time cost and estimation accuracy.
% Note that the estimation accuracy is measured under Overlap@$k$,
% which indicates the similarity of detected barrier words under different
% budgets compared with the exact case.
% Figure~\ref{fig:overlap-at-k} shows that using budget around 50 or 100
% can reach the ground truth risk rank obtained by exactly calculation
% over all $\vert \mathcal{V}^s \vert$ edited inputs per position, while
% keeps a relative acceptable speed.

% \begin{figure*}[t]
%     \centering
%     \includegraphics[scale=0.52]{fig/box_plots.pdf}
%     \caption{The Overlap@$k$ of different budgets on different samples; for each budget, the concentrated mean can be seen as the average overlap rate of that budget.}
%     \label{fig:box-plots}
% \end{figure*}

% Here we propose to use unbiased or biased
% Monte Carlo methods to estimate the truncated mean
% in Eq.~\eqref{eq:truncated-mean}.
% Specifically, we assume a budget $b$ based on which we
% conduct real decoding to obtain $b$ metric values
% $\mathcal{S}^b \subset \mathcal{S}$.

\noindent
\textbf{Uniform}
A very simple unbiased estimator of $tm(\tn{x}_i)$ is to uniformly $b$ elements from $\textrm{Edit}(\tn{x},i)$, and compute the mean of those $m$s that are larger than $m^o$.
% However, since the number of effective budget $b^{>m^o}$ can be much
% smaller than $b$ which can potentially lead to large variance of
% the estimator, we try other two biased sampling methods which
% can potentially increase $b^{>m^o}$.
However, since we do not restrict the substitutions, two
potential issues might lead to large variance of uniform sampling:
a) waste of budget: substitutions that lead
to metric values lower than $m^o$ could be more;
b) hardness of coverage (less concentrated): wider the range of the
orange band (in the histogram of
Figure~\ref{fig:example-risk-dist}), larger the variance.

\noindent
\textbf{Stratified}
To be less stochastic to combat variance,
we can first use uniform sampling
for randomly picking $B$ elements from $\textrm{Edit}(\tn{x},i)$, and then
use the loss function
\begin{equation}
  \mathcal{L}_{\hat{\theta}}(\tilde{\tn{x}}) := - \log P(\tn{y} \vert \tilde{\tn{x}}; \hat{\theta})
  \label{eq:str-loss}
\end{equation}
as a surrogate to choose the top-$b$ from the $B$ choices.
The first stage respects the uniform
distribution in $\textrm{Edit}(\tn{x},i)$, while the second stage is deterministic (i.e., top-$b$ likelihood values) which
can potentially lower the variance.
% Since computing the loss takes less time
% than real decoding, we can first uniformly sample a much larger budget $B$
% to calculate $B$ losses, based on which to choose the top-$b$
% edit choices from $B$. The second level of the stratified strategy
% is deterministic which can potentially lower the variance.

\noindent
\textbf{Gradient-aware}
To avoid the sampling budget hyper-parameter $B$ at the first stage
of the stratified method, we can utilize the gradient
of the original loss $\mathcal{L}_{\hat{\theta}}(\tn{x})$ which guides
the change of embeddings of $\tn{x}_i$ that can minimize the loss:
\begin{equation}
    \text{Emb}'(\tn{x}_i) = \text{Emb}(\tn{x}_i) - 1.0 \cdot \nabla_{\text{Emb}(\tn{x}_i)} \mathcal{L}_{\hat{\theta}}(\tn{x}).
\label{eq:one-step-grad}
\end{equation}
\noindent
Contrary to the method of adversarially modifying the input
in \citet{Cheng2019RobustNM},
we conduct 1-step gradient update with
learning rate 1.0 to \textit{minimize} the original loss,
and then use the normalized dot product similarity between the
updated embedding and all other embeddings of the source vocabulary
to bias the sampling of $b$ elements from $\textrm{Edit}(\tn{x},i)$.

The entire algorithmic procedures of the three estimators
are summarized succinctly in Algorithm~\ref{alg:eval-risk}.

% \subsubsection{Comparing the estimators}

% % Mean performance result on train and dev/test set
% \begin{table*}[t]
% \setlength{\tabcolsep}{4.0pt}
% \centering
% \small{
% \begin{tabular}
% { c||c||c|c|c|c|c|c|c }
% \hline
% \textbf{task} & \textbf{model} & \textbf{train} & \textbf{dev 02} & \textbf{03} & \textbf{04} & \textbf{05} & \textbf{06} & \textbf{08} \\ \hline \hline
% \multirow{3}{4em}{Zh$\Rightarrow$En} & rnn & 35.02 & 41.02 & 38.76 & 41.75 & 38.07 &38.91 & 31.20 \\ 
% & fconv & 40.02 & 45.58 & 44.18 & 46.15 & 43.85 & 44.68 & 36.34 \\ 
% & san   & 38.46 & 47.85 & 46.20 & 47.66 & 46.52 & 46.73 & 38.75 \\ \hline \hline
% \multirow{3}{4em}{En$\Rightarrow$Zh} & rnn & 39.12 & 22.57 & 16.08 & 18.39 & 15.50 & 19.05 & 14.07 \\ 
% & fconv & 40.91 & 24.96 & 17.81 & 20.22 & 17.07 & 21.08 & 15.99 \\ 
% & san   & 41.67 & 26.31 & 19.44 & 21.46 & 18.08 & 22.27 & 16.26 \\ 
% \end{tabular}
% }
% \caption{The main generalization performance results.}
% \label{tab:main-results}
% \end{table*}

\section{Experimental Conditions}
\label{sec:experiment-condition}

% % Mean performance result on train and dev/test set
% \begin{table}[t]
% \setlength{\tabcolsep}{4.0pt}
% \centering
% \small{
% \begin{tabular}
% { c||c||c|c|c}
% \textbf{Task} & \textbf{Model} & \textbf{Train} & \textbf{Dev.} & \textbf{Test Avg.} \\ \hline \hline
% \multirow{3}{4em}{Zh$\Rightarrow$En} & rnn & 35.02 & 41.02 & 37.73 \\ 
% & fconv & 40.02 & 45.58 & 43.04 \\ 
% & san   & 38.46 & 47.85 & 45.17 \\ \hline \hline
% \multirow{3}{4em}{En$\Rightarrow$Zh} & rnn & 39.12 & 22.57 & 16.61 \\ 
% & fconv & 40.91 & 24.96 & 18.43 \\ 
% & san   & 41.67 & 26.31 & 19.50 \\  \hline
% \end{tabular}
% }
% \caption{The main generalization performance results on NIST benchmark measured by BLEU; note that here \textbf{Train} is measured through single reference while \textbf{Dev.} is measured by four references for the Zh$\Rightarrow$En task, so for rnn, \textbf{Dev.} can surpass \textbf{Train}.}
% \label{tab:model-performance}
% \end{table}

In this section, we set up the overall experimental
scenarios regarding the data configuration
and the model architectural choice.

\noindent
\textbf{Data settings}
We conduct experiments on Zh$\Rightarrow$En and En$\Rightarrow$Zh
translation tasks using the well-known NIST benchmark.
The dev and test datasets of the NIST benchmark
are marked by year, e.g. NIST02 (dev), NIST03 etc. For
Zh$\Rightarrow$En, each dev/test
source sentence has four
references; and for En$\Rightarrow$Zh,
we pick the first source input of the four as the
source-side instance. During truncated mean estimation
stage, for the Zh$\Rightarrow$En translation task, we use the
first reference as the ground truth in smoothed
sentence-level BLEU calculation.
% To cover the whole spectrum of the vocabulary, we follow the following rules to randomly choose certain instances according to word frequency: a) for word with $\text{freq} > 50$, randomly sample 5 instances containing the word; b) with $10 < \text{freq} \leq 50$, randomly sample 1 instance; c) with $\text{freq} \leq 10$, no instance is sampled. We conduct such train set instance selection for each translation task respectively since their vocabularies differ. And the resulting train/dev/test/unseen splits are shown in Table~\ref{tab:data-summary}.

\noindent
\textbf{Model settings} We consider three types of basic model architectures proposed in
\citet{Luong2015StanfordNM,gehring2017convolutional,vaswani2017attention} respectively,
representing the advancement of architectural inductive bias in recent years.
% The number of parameters and number of layers of each model architecture are shown in Table~\ref{tab:model-spec}.
Their average performance over NIST03, 04, 05, 06, 08 are
summarized in Table~\ref{tab:model-performance}
in Appendix~\ref{sec:overal-mean-performance}.

\section{Analyses}

\subsection{Comparing the estimators~\footnote{More detailed informations about the evaluation metrics used in this subsection are in Appendix~\ref{sec:eval-metrics}.}}

We conduct simulation experiments among 50
unseen sentence pairs from NIST03
with whole vocabulary decoding
to compute the ground truth truncated mean for each $\tn{x}_i$
with Eq.~\eqref{eq:truncated-mean},
and then compare the above proposed sampling methods in terms of
\textit{overlap@$k$}, \textit{variance} or \textit{rank stability}
of the estimator under different budgets
$b=5, 10, 25, 50, 100, 250, 500, 1000, 5000$.
For the stratified strategy, we set $B$ to 500 for $b<500$
budgets, $B=1000$ for $b=500$, $B=2000$ for $b=1000$,
and $B=10000$ for $5000$.
To be statistically significant,
for each source word $\tn{x}_i$, we repeat the estimation
procedure for $r=25$ times.

\begin{figure}[ht]
    \centering
    \includegraphics[scale=0.55]{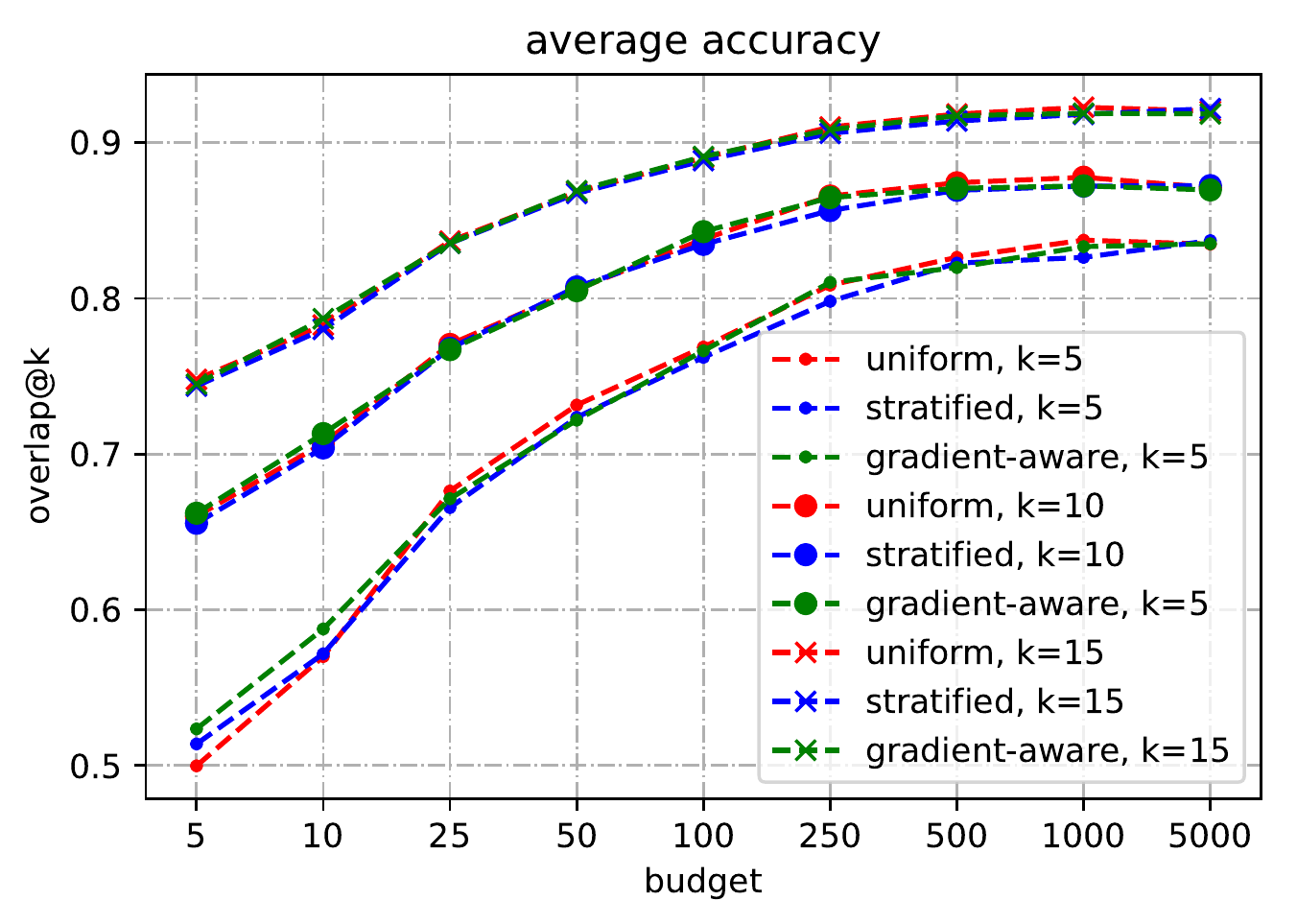}
    \caption{The overlap@$k$ metric values over the three proposed estimation methods on the 50 samples under different budgets (5 to 5000); $k$ is set to 5, 10 and 15.}
    \label{fig:avg-acc}
\end{figure}

\begin{figure}[ht]
    \centering
    \includegraphics[scale=0.55]{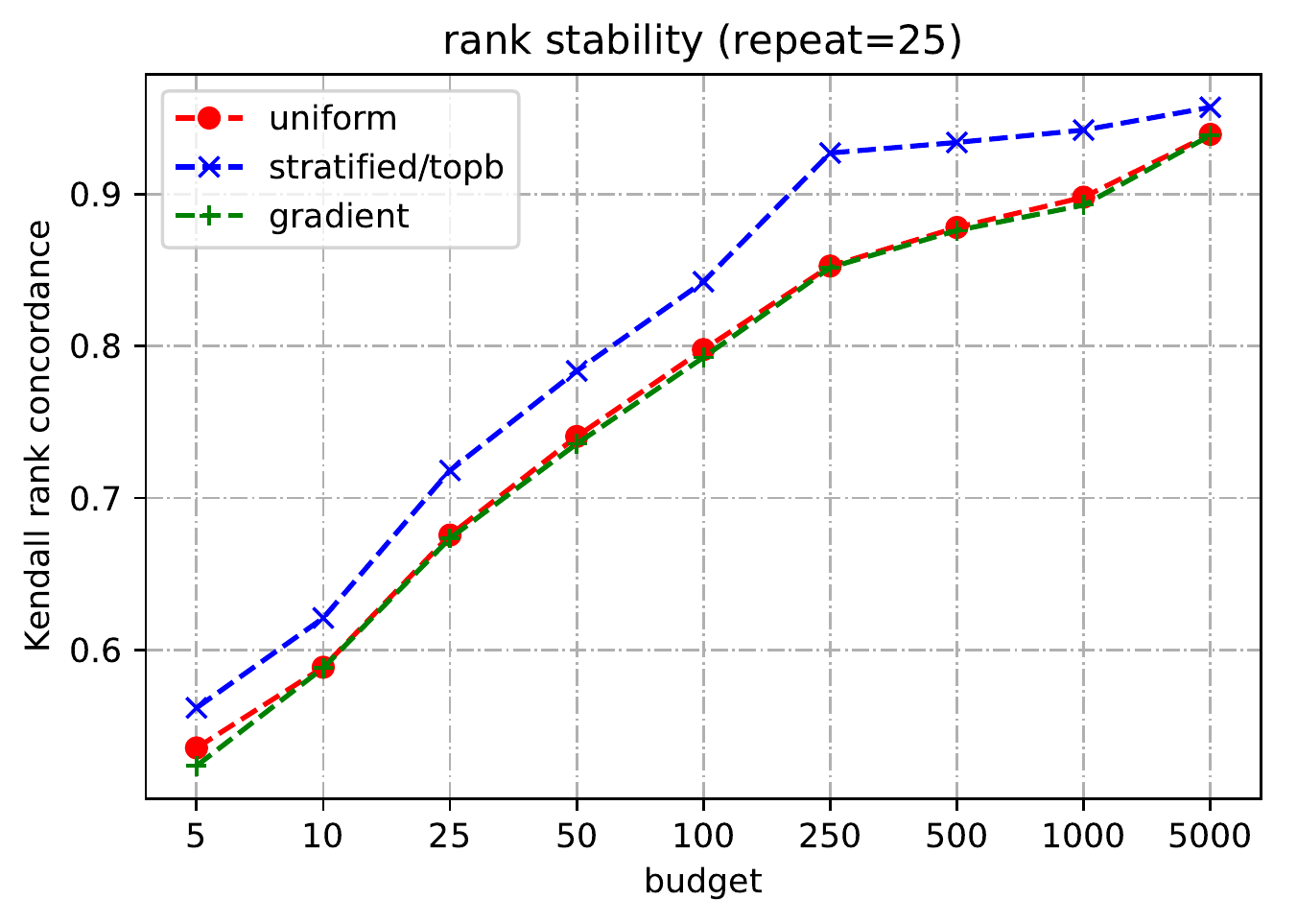}
    \caption{The rank stability of the three proposed estimation methods under different budgets. They are averaged over the 50 chosen samples and measures the variance of methods over $25$ repeated experiments.}
    \label{fig:rank-stability}
\end{figure}

% \begin{table}
%     \setlength{\tabcolsep}{2.0pt}
%     \centering
%     \small{
%     \begin{tabular}{c||ccccc|cccc}
%                     & \multicolumn{5}{c|}{\textit{sec.}} & \multicolumn{4}{c}{\textit{min.}} \\ \hline \hline
%         budget $b$  & 5 & 10 & 25 & 50 & 100 & 0.25\textit{k} & 0.5\textit{k} & 1\textit{k}  & 5\textit{k} \\ \hline
%         time cost   & 4 & 7  & 17 & 33 & 65  & 3 & 6 & 11 & 56
%     \end{tabular}
%     }
%     \caption{Time complexity for the uniform estimator among different budgets; note that the time cost is an average measure over each sentence.}
%     \label{tab:time-cost}
% \end{table}

\begin{table}[t]
    \setlength{\tabcolsep}{2.0pt}
    \centering
    \small{
    \begin{tabular}{c||ccccc|c|c|c}
        \hline
                    & \multicolumn{8}{c}{\textit{second per sentence}}  \\ \hline \hline
        budget $b$  & 5 & 10 & 25 & 50 & 100 & 250 & 500 & 1000 \\ \hline
        time cost   & 4 & 7  & 17 & 33 & 65  & 180 & 360 & $>$ 600 \\ \hline
    \end{tabular}
    }
    \caption{The time complexity for the uniform estimator among different budgets; note that the time cost is an average measure over each sentence.}
    \label{tab:time-cost}
\end{table}

\noindent
\textbf{Accuracy}
We use the overlap@$k$ metric to measure the similarity between top-$k$
risky words with exact and approximate risk calculation methods.
As demonstrated in Figure~\ref{fig:avg-acc}, different methods
lead to very overlapped performance. And with a budget larger than 100,
it can lead to an average overlap@$k$ around 85\%, based on
which we think is enough for the subsequent analyses.

\noindent
\textbf{Variance}
The rank stability is measured through Kendall's coefficient
of concordance~\citep{Mazurek2011EVALUATIONOR} which essentially
calculates the similarity among different (repeat=25) ranks of
each
sample.
The larger the value is, the more consistent among
different runs the ranks stay, thus smaller variance of the estimator.
As shown in Figure~\ref{fig:rank-stability},
the uniform and gradient-aware estimators have similar variance while
the stratified estimator has lower variance, which might be benefited
from its second deterministic stage.

\noindent
\textbf{Complexity}
We also summarize the time cost of each budget $b$ in
Table~\ref{tab:time-cost}. Since most of the time complexity
comes from real decoding, here we only measure the
time cost of the uniform estimator.
We test the process on a single M40 GPU.

As a trade-off between accuracy, variance and time complexity,
we adopt the stratified strategy with budget $B=500, b=100$ as our
approximate detection method in all subsequent analyses.
This takes around 16 hours for 1\textit{k} sentences with a
decent detection accuracy around 85\% with respect to overlap@$k$
and nice rank stability up to 84\%.

\subsection{Characterizing the Generalization Barrier Words}
\label{sec:characterize}

In this section, we try to characterize
the detected barrier words from different perspectives, i.e,
to understand them with their linguistic properties and
their comparison with respect to other source word categorizations
in statistical senses.
~\footnote{We use the great toolkits LTP~\citep{che-etal-2010-ltp}
and AllenNLP~\citep{gardner-etal-2018-allennlp} for the basic
linguistic analyses of Chinese and English respectively.}
% \begin{itemize}
%     \item Do the detected barrier words mostly fall into a subword (BPE) segment? And what is the distribution of barrier words with respect to part-of-speech (POS) categories?

%     \item Are those detected barrier words cross-model stable locally and globally?
% \end{itemize}

\subsubsection{Linguistic properties}

\begin{table}
\setlength{\tabcolsep}{2.8pt}
  \begin{subtable}[t]{1.\linewidth}
    \centering%
    \small{
    \begin{tabular}{c||l|l|l||l}
    \hline
    \textbf{POS cat.} & \textbf{k=5} & \textbf{k=10} & \textbf{k=15} & \textbf{base} \\ \hline \hline
    BPE      & 15.04\%$^-$  & 14.88\%$^-$ & 14.72\%$^-$ & 15.33\% \\
    Noun     & 16.22\% $^-$ & 16.13\%$^-$ & 16.48\%$^-$ & 17.63\% \\
    \dashuline{Prop. N.} &  \dashuline{6.19\%$^-$}  &  \dashuline{6.57\%$^-$} &  \dashuline{6.51\%$^-$} &  \dashuline{7.44\%} \\
    Pron.    &  1.79\% $^-$ &  2.15\%$^-$ &  2.36\%$^+$ &  2.35\% \\
    Verb     & 18.81\%$^+$  & 18.73\%$^+$ & 18.93\%$^+$ & 18.36\% \\
    Adj.     & 2.54\%$^-$   & 2.84\%$^-$  &  2.93\%$^-$ & 3.19\%  \\
    Adv.     & 4.49\%$^+$   & 4.35\%$^+$  &  4.32\%$^+$ &  4.07\%  \\
    \uwave{Prep.}    &  \uwave{4.42\%$^+$}  &  \uwave{4.33\%$^+$} &  \uwave{4.43\%$^+$} & \uwave{3.83\%}  \\
    \uwave{Punc.}    & \uwave{15.24\%$^+$}  & \uwave{14.11\%$^+$} & \uwave{13.37\%$^+$} & \uwave{11.44\%} \\
    Q\&M     &  4.24\%$^-$  &  4.74\%$^-$ &  4.65\%$^-$ & 4.87\% \\
    C\&C     &  1.58\%$^-$  &  2.02\%$^-$ &  2.06\%$^-$ & 2.23\% \\ \hline
    \end{tabular}
    \caption{on NIST03 Zh$\Rightarrow$En direction}\label{tab:pos-distribution}
    }
  \end{subtable}\par\bigskip

  \begin{subtable}[t]{1.\linewidth}%
    \centering%
    \small{
    \begin{tabular}{c||l|l|l||l}
    \hline
    \textbf{POS cat.} & \textbf{k=5} & \textbf{k=10} & \textbf{k=15} & \textbf{base} \\ \hline \hline
    \dashuline{BPE}      & \dashuline{10.18\%$^-$}      & \dashuline{10.86\%$^-$} & \dashuline{11.33\%$^-$} & \dashuline{12.00\%}  \\
    \dashuline{Noun}     & \dashuline{21.90\%$^-$}      & \dashuline{22.71\%$^-$} & \dashuline{22.44\%$^-$} & \dashuline{24.07\%}  \\
    Pron.    & 2.17\%$^+$       &  2.26\%$^+$ &  2.30\%$^+$ &  2.15\%  \\
    Verb     & 11.98\%$^+$      & 11.41\%$^+$ & 11.66\%$^+$ & 11.26\%  \\
    Adj.     &  7.15\%$^-$      &  7.43\%$^-$ &  7.67\%$^-$ &  8.19\%  \\
    Adv.     &  3.14\%$^+$      &  3.07\%$^+$ &  3.06\%$^+$ &  2.93\%  \\
    \uwave{Prep.}    & \uwave{13.13\%$^+$}      & \uwave{13.10\%$^+$} & \uwave{12.74\%$^+$} & \uwave{11.88\%}  \\
    \uwave{Punc.}    & \uwave{14.64\%$^+$}      & \uwave{13.03\%$^+$} & \uwave{12.33\%$^+$} & \uwave{10.41\%}  \\
    Det.     & 8.39\%$^-$       &  8.91\%$^-$ &  9.20\%$^+$ &  9.05\%  \\
    C\&C     & 1.74\%$^-$       &  1.81\%$^-$ &  1.86\%$^-$ &  2.20\% \\ \hline
    \end{tabular}
    }
    \caption{on NIST03 En$\Rightarrow$Zh direction}\label{tab:pos-distribution-enzh}
  \end{subtable}%
%   \caption{Distribution of the detected generalization barrier words according to the Chinese~\citep{che-etal-2010-ltp} and English~\citep{gardner-etal-2018-allennlp} POS category.}
    \caption{Distribution of the detected generalization barrier words according to Part-of-Speech category.}
  \label{tab:pos-distribution-all}
\end{table}

\noindent
\textbf{Distribution over Part-of-Speech}
In this part, we summarize the distribution of the
detected generalization barrier words with respect to
their Part-of-Speech (POS) tags.
In order to consider the subword segments, we first
use a POS tagger to label on the BPE-restored corpus,
and then map the non-subword
segments to the corresponding POS tags
while the subword segments to a special tag
named BPE, so that we can readily measure the ratio of subwords.
The summary statistics are shown in
Table~\ref{tab:pos-distribution-all}.
To compare with the natural distribution of all the words over POS,
we also demonstrate them together with the detected
generalization barrier words at the \textbf{base} column.

For both Chinese and English source inputs,
barrier words are pervasive across all POS categories, since
there is no significant difference from the base distribution.
Note that, functional words like preposition and punctuation increase
the most (with 3 $^+$) over the base. For English source, BPE
is less tended to be barriers which indicate the benefit of subword-based
segmentation. And for content words like noun and proper noun, they tend to
be relatively less ambiguous and less context dependent thus tend to cause
less problems.

% For Zh$\Rightarrow$En translation,
% the conclusion is that the detected
% generalization barrier words come from all categories
% of POS, which almost follows the natural distribution
% (similar to the \textbf{base} column). However, words
% from the preposition category or the verb category
% seems to slightly increase (3$^+$) or decrease (3$^-$),
% which indicates their relatively larger or smaller
% tendency of being the generalization barrier words.

% \subsubsection{Branches or leaves?}

% \begin{table}[t]
%     \setlength{\tabcolsep}{3.5pt}
%     \centering
%     \begin{tabular}{c||c|c|c|c}
%     \textbf{Sem. role} & \textbf{k=5} & \textbf{k=10} & \textbf{k=20} & \textbf{base} \\ \hline \hline
%     Sub.      &  &  & \\
%     Obj.     &  &  & \\
%     Pred.     &  &  & \\
%     Others     &  &  & \\ \hline
%     \end{tabular}
%     \caption{Distribution of the detected generalization barrier words according to POS tags.}
%     \label{tab:slr-distribution}
% \end{table}
\begin{table}[t]
    \setlength{\tabcolsep}{2.8pt}
    \centering
    \small{
    \begin{tabular}{c||l|l|l||l}
    \hline
    \textbf{Distance} & \textbf{k=5} & \textbf{k=10} & \textbf{k=15} & \textbf{base} \\ \hline \hline
    1      &  54.49\%$^+$  & 54.08\%$^+$  & 53.40\%$^+$  & 51.17\%  \\
    2      &  85.31\%$^+$  & 85.44\%$^+$  & 85.20\%$^+$  & 83.75\%  \\ \hline
    %3      &       &  &  & 11.26\%  \\
    %4      &       &  &  & 2.20\% \\ \hline
    \end{tabular}
    }
    \caption{The recall@$(k, d)$ statistics with respect to the distance $d=(1, 2)$ to all the leaves on the dependency tree (on NIST03 Zh$\Rightarrow$En direction).}
    \label{tab:deptree-distribution}
\end{table}

% \begin{table}[t]
%     \setlength{\tabcolsep}{2.8pt}
%     \centering
%     \small{
%     \begin{tabular}{c||l|l|l||l}
%     \textbf{Distance} & \textbf{k=5} & \textbf{k=10} & \textbf{k=15} & \textbf{base} \\ \hline \hline
%     1      &       &  &  & 12.00\%  \\
%     2      &       &  &  & 24.07\%  \\ \hline
%     %3      &       &  &  & 11.26\%  \\
%     %4      &       &  &  & 2.20\% \\ \hline
%     \end{tabular}
%     }
%     \caption{The recall@$(k, d)$ statistics with respect to the distance to all the leaves on the dependency tree (on NIST03 En$\Rightarrow$Zh direction).}
%     \label{tab:deptree-distribution-enzh}
% \end{table}

\noindent
\textbf{Branch or leaves?}
% In this part, we conduct an analysis to shed light on the question that:
% do the barrier words mostly come from main branch of an input sentence
% which may mainly reflect the core semantics of the sentence, or from
% the attachment structure that modifies the main semantics to be fine-grained?
% To achieve this, we take advantage of off-the-shelf dependency parser
% to label the dependency structure of the input sentence. Then we draw
% the distribution of overlap metric value with respect to the distance
% towards the root or the leaves of the dependency tree. This analysis
% can to some extent reflect the degree of human alignment of current
% state-of-the-art NMT systems, since the branch meaning of a sentence
% should be more reliably conveyed in certain situation than leaves information.
% Table~\ref{tab:deptree-distribution} summarizes the statistics.
In this part, we conduct an anlysis to shed light on the question that:
do barrier words mostly come from main branch
of the source dependency tree or modifiers?
Since AllenNLP's dependency parser re-tokenizes the
original sources for English, we only provide
statistics on Chinese in Table~\ref{tab:deptree-distribution}. Distance means
distance of each word towards the leaves of the dependency tree, the
\textbf{base} column also shows how much
words are covered under certain distance. And
other entries with specific $k$ means
how much top-$k$ risky words (the detected barrier words)
are recalled. There is a little tendency that
barrier words tend to be more close to the leaves than branch.

% We also conduct semantic role labeling of each input sentence
% to divide the source words
% into two parts with the same philosophy as dependency tree and compute
% the occupation ratio of the barrier words with respect to these two parts.
% Table~\ref{tab:slr-distribution} summarizes the statistics.
% We also conduct constituent parsing of each input sentence to divide
% the source words into two parts: shallow recursive and deep recursive
% in the same philosophy of dependency tree.

\subsubsection{Comparing to other source word categorizations}
\label{sec:compare-with-other-source-cat}

\begin{table}
\setlength{\tabcolsep}{2.8pt}
    \centering
    \small{
    \begin{tabular}{c|c||l|l|l}
    \hline
    \textbf{Task} & \textbf{Word cat.} & \textbf{k=5} & \textbf{k=10} & \textbf{k=15}\\ \hline \hline
    \multirow{4}{*}{Zh$\Rightarrow$En} & Random         & 21.16\%      & 40.79\%   & 58.09\%  \\ \cdashline{2-5}
    & Frequency      & 20.83\%      & 40.23\%   & 55.62\%  \\
    & Entropy        & 22.17\%      & 42.81\%   & 58.51\%  \\
    & Exception      & 21.75\%      & 42.98\%   & 58.16\%  \\ \hline
    \multirow{4}{*}{En$\Rightarrow$Zh} & Random         & 18.90\% & 36.90\%  & 52.27\%  \\ \cdashline{2-5}
    & Frequency      & 18.24\% & 34.86\%  & 50.11\%  \\
    & Entropy        & 20.76\% & 38.17\%  & 53.28\%  \\
    & Exception      & 18.47\% & 36.83\%  & 51.89\%  \\ \hline
    \end{tabular}
    }
  \caption{The overlap@$k$ statistics with respect to different types of troublesome word statistics methods which due not utilize real decoding.}\label{tab:fast-detection-method-overlap-all}
\end{table}

In this part, we compare the detected generalization barrier words with other
source word categorizations: a) low-frequency words; b) high translation
entropy words~\citep{zhao2019addressing}; and
c) exception words~\citep{zhao2018addressing}. They are all based on certain
global statistical clues of the training corpus, i.e. alignments
obtained \textit{extrinsically}.
% The low-frequency words are usually said to be hard to translate,
Words in a) are commonly said to cause generalization error, while
words in b) and c) are dubbed as under-translated and troublesome words
respectively according to the papers.
Here, we want to know whether those probable trouble makers
are generalization barrier words?

% At least two recent works~\citep{zhao2018addressing,zhao2019addressing}
% have started to focus on troublesome words or under-translated words
% of NMT and try to resolve them respectively with specific solutions.
% Here, instead, we want to know whether those troublesome words
% or under-translated words are generalization barrier words?
% If so, we can use their proposed heuristics for accurate
% detection of those generalization barrier words.

Since a) - c) all use global statistics for each word $v \in \mathcal{V}$,
to compare with the generalization barriers annotated with local risk,
for each unseen input $\tn{x}$, we also use $\tn{x}_i$'s global statistical
clue to annotate itself in this local context so that overlap@$k$ can
be used for comparison. The statistics are denoted as
$1/freq(v)$, $te(v)$, $er(v)$ for inverse frequency, translation entropy
and exception rate.
Translation entropy of $v$ is obtained through estimating the lexical
translation probability $\phi(w \vert v)$ and compute the entropy of this
distribution among all $w \in \mathcal{V}'$ (target vocabulary). Exception rate of a word $v$
is calculated through the ratio between the number of exception alignment
according to certain exception condition and the total number of alignment
of $v$ across the training corpus,
$\frac{M^v}{N^v}$. Detailed introduction of the trouble makers is in Appendix~\ref{sec:detailed-info-for-trouble-makers}.
% ~\footnote{All alignments are obtained through
% $\text{fast\_align}$~\citep{dyer-etal-2013-simple},
% detailed definitions are shown
% in Appendix xxx.}

Table~\ref{tab:fast-detection-method-overlap-all} shows the overlap@$k$
values for Zh$\Rightarrow$En and En$\Rightarrow$Zh. The \textbf{random}
row shows the metric values if we randomly choose an order of the source
words. It is obvious that all categorizations are very close to random,
with Entropy slightly better than random,
which indicates our generalization barrier words that rely statistics
from inference-aware counterfactuals are very different.
This highlights the novelty of such phenomenon,
and implies the importance of studying generalization with
explicit inference under consideration.

\subsection{Context-sensitive/agnostic barriers}

\begin{figure}[!h]
    \centering
    \includegraphics[scale=0.4]{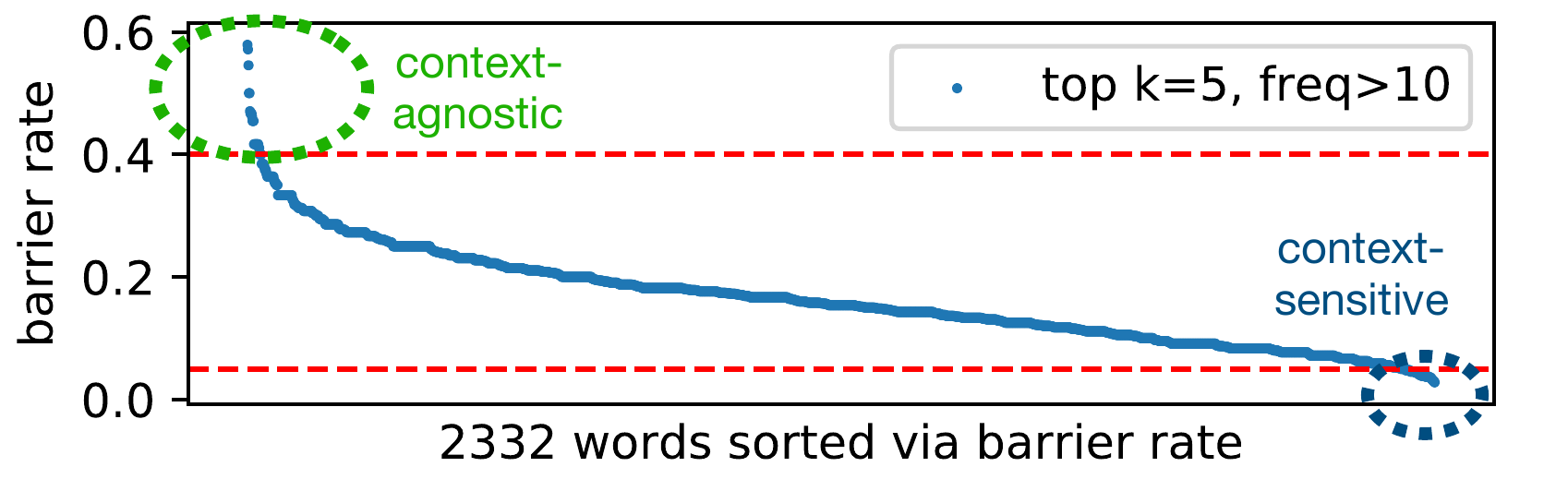}
    \caption{The distribution of barrier rate across words with context count larger than 10 on NIST 03-06.}
    \label{fig:barrier-rate}
\end{figure}

In this section, we try to aggregate local statistics
to obtain certain global understanding: is it possible that
some words are prone to be generalization barriers in
a context-agnostic way or the reverse.
We aggregate the top-$k$ words in each test input
and calculate their count. Specifically, if one appearance
of a word roughly
represents a context, we can calculate the probability of certain
detected barrier word of being an universal barrier according to
the following barrier rate:
% \begin{equation}
%     p(v) = \frac{\sum_i \text{is\_Barrier}[v \vert v \in \tn{x}^i]}{\sum_i \mathbb{I}[v | v \in \tn{x}^i]}
% \end{equation}
\begin{equation}
    p(v) = \frac{\sum_i \text{Count}\Big(v \vert \text{is\_Barrier}(v) \wedge v \in \tn{x}^i\Big)}{\sum_i\text{Count}\Big(v | v \in \tn{x}^i\Big)}
\end{equation}
We then summarize the distribution of each word's barrier rate
in Figure~\ref{fig:barrier-rate}. The two horizontal dashed red
lines are 0.4 and 0.05, indicating highly context-agnostic and
context-sensitive respectively. As you can see, there are
few context-agnostic barriers and most of the barrier words
are very sensitive to context, indicating the necessary
of pursuing large-scale training data with
abundant contexts~\citep{schwenk2019wikimatrix,schwenk2019ccmatrix}.

\subsection{Complementary across architectures}

% \begin{table}
% \setlength{\tabcolsep}{2.8pt}
%   \begin{subtable}[t]{1.\linewidth}%
%     \centering%
%     \small{
%     \begin{tabular}{c||l|l|l}
%     \textbf{Architecture pair} & \textbf{k=5} & \textbf{k=10} & \textbf{k=15}\\ \hline \hline
%     san-fconv         & 28.65\%  & 45.03\%  & 58.42\%  \\
%     san-rnn           & 25.46\%  & 43.73\%  & 57.68\%  \\
%     fconv-rnn         & 27.64\%  & 45.52\%  & 58.85\%  \\ \hline
%     \end{tabular}
%     }
%     \caption{on NIST03 Zh$\Rightarrow$EN direction}\label{tab:architectural-overlap}
%   \end{subtable}%
%   \par\bigskip

%   \begin{subtable}[t]{1.\linewidth}%
%     \centering%
%     \small{
%     \begin{tabular}{c||l|l|l}
%     \textbf{Architecture pair} & \textbf{k=5} & \textbf{k=10} & \textbf{k=15}\\ \hline \hline
%     san-fconv         & 24.13\%  & 39.62\%  & 53.18\%  \\
%     san-rnn           & 24.80\%  & 40.40\%  & 53.41\%  \\
%     fconv-rnn         & 23.70\%  & 40.11\%  & 53.24\%  \\ \hline
%     \end{tabular}
%     }
%     \caption{on NIST03 En$\Rightarrow$Zh direction}\label{tab:architectural-overlap-enzh}
%   \end{subtable}%

%   \caption{The overlap@$k$ statistics with respect to different architectural choices.}\label{tab:architectural-overlap-all}
% \end{table}

\begin{table}
\setlength{\tabcolsep}{2.8pt}
    \centering
    \small{
    \begin{tabular}{c|c||l|l|l}
    \hline
    \textbf{Task} & \textbf{Arch. pair} & \textbf{k=5} & \textbf{k=10} & \textbf{k=15}\\ \hline \hline
    \multirow{3}{*}{Zh$\Rightarrow$En} & san-fconv         & 28.65\%  & 45.03\%  & 58.42\%  \\
    & san-rnn           & 25.46\%  & 43.73\%  & 57.68\%  \\
    & fconv-rnn         & 27.64\%  & 45.52\%  & 58.85\%  \\ \hline
    \multirow{3}{*}{En$\Rightarrow$Zh} & san-fconv         & 24.13\%  & 39.62\%  & 53.18\%  \\
    & san-rnn           & 24.80\%  & 40.40\%  & 53.41\%  \\
    & fconv-rnn         & 23.70\%  & 40.11\%  & 53.24\%  \\ \hline
    \end{tabular}
    }
  \caption{The overlap@$k$ statistics with respect to different architectural choices (pair-wise comparison).}\label{tab:architectural-overlap-all}
\end{table}

On the same training corpus,
we train three different model architectures mentioned
in Section~\ref{sec:experiment-condition} based on
\textit{rnn}, \textit{fconv} and \textit{san} respectively.
To measure the similarity of the detected generalization barrier
words between every two of them, we also use the overlap@$k$ metric.
Their pair-wise overlap statistics are shown in
Table~\ref{tab:architectural-overlap-all}. It seems that the
detected barrier words are very sensitive to architectural choice
which might indicate that combining the best-practice architectures
through ensemble methods might be a method for alleviating barriers
specific to certain architecture.

% \subsubsection{Training data choice}

% We divide the training corpus into five splits [1, 2, 3, 4, 5]
% with identical sizes, and conduct training of the same architecture
% (\textit{san}) on the five corpora with certain split left out respectively.
% Then we compute overlap@$k$ metric for any two pairs.

% In this subsection, we try to unveil the causal effect of barrier words by answering
% the following questions:
% \begin{itemize}
%     \item How much overall performance gain can we obtain by editing the barrier words?
%     \item Does the performance gain differ from the gain obtained by hypotheses re-ranking?
% \end{itemize}
% \subsection{Understanding detected barrier words}

% \subsubsection{Characterization}

% In this section, we try to characterize some properties of
% the detected barrier words by answering the following questions:
% \begin{itemize}
%     \item Do the detected barrier words mostly fall into a subword (BPE) segment?
%     \item What is the distribution barrier words with respect to part-of-speech categories?
%     \item Does the performance gain differ from the gain obtained by hypotheses re-ranking?
%     \item Are those detected barrier words cross-model stable locally and globally?
% \end{itemize}

% \subsubsection{Causal analysis}

\subsection{Potential usage of the barrier words}
\label{sec:potential-usage}

After obtaining an understanding of the generalization barrier words,
in this part, we try to present one potential usage of them which
we think
could be more relevant to improving the translation performance of NMT
systems in an automatic way through re-ranking~\citep{yee-etal-2019-simple}.
We show that in Table~\ref{tab:reranking-compare-all}
by editing the barrier words randomly to form
a groups of inputs, we can generate a collection of re-ranking hypotheses
with higher oracle performance, better translation recall and diversity
than top-$k$ candidates from one single input, which is currently the
common wisdom of re-ranking for NMT, i.e. using top-$k$ scored beam search
candidates as outputs. Actually, we find that, the usual
top-$k$ candidates are very similar to each other and the oracle translation seems
to be a paraphrased version of the highest model-scored one which might be very hard
for the reranking model to pick up, instead the candidates generated
by editing barriers can recall the actual incorrectly or un-translated parts of
meaning of the source.
Details for the measures used here are introduced in
Appendix~\ref{sec:metrics-for-reranking}.

\begin{table}
\setlength{\tabcolsep}{2.3pt}
    \centering%
    \small{
    \begin{tabular}{c|c||l|l|l}
    \hline
    \textbf{Task} & \textbf{Candidate} & \textbf{Oracle}  $\uparrow$ & \textbf{Coverage} $\uparrow$ & \textbf{Diversity} $\downarrow$ \\ \hline \hline
    \multirow{2}{*}{Zh$\Rightarrow$En} & top-$k$        & 39.40  & 78.22  & 61.78  \\
    & barrier        & 42.78 \tiny{\color{red}{(+3.38)}}  & 83.88 \tiny{\color{red}{(+5.66)}}  & 57.21 \tiny{\color{red}{(-4.75)}}  \\ \hline
    \multirow{2}{*}{En$\Rightarrow$Zh} & top-$k$     & 32.31  & 72.10  & 59.98  \\
    & barrier     & 37.48 \tiny{\color{red}{(+5.17)}} & 79.62 \tiny{\color{red}{(+7.52)}}  & 52.72 \tiny{\color{red}{(-7.26)}}  \\ \hline
    \end{tabular}
    }
  \caption{The comparison of various properties of the re-ranking candidates generated through the traditional top-$k$ and our barrier-editing methods.}\label{tab:reranking-compare-all}
\end{table}

% \section{Discussion}

\section{Conclusion and Future Work}

In this paper, we identify and define a new phenomenon
in NMT named generalization barriers through inference-aware
counterfactual analyses.
Simple approximate methods are investigated to
better detect such generalization barrier words.
After large-scale detection on held-out test sets, we
find that barrier words are pervasive among different
POS categories and mostly prone to be functional words.
However, they are very different from previous identified
trouble makers in the source side. Moreover, barrier
words are tend to be more context-sensitive and less universal.
We can potentially alleviate them through ensembles of different
architectures~\citep{Athiwaratkun2019ThereAM}
or editing them for constructing better
re-ranking candidates~\citep{yee-etal-2019-simple}.
Future work involves fundamental causal
analysis of the emergence of
such phenomenon \textit{intrinsically}
through the lens of the learned representation
and representation confounding effect~\citep{li2019approximate}
or \textit{extrinsically} through
compositionality study of the input.

% \section*{Acknowledgments}

% The acknowledgments should go immediately before the references. Do not number the acknowledgments section.
% Do not include this section when submitting your paper for review.

% \bibliography{anthology}
\bibliography{customized}
\bibliographystyle{acl_natbib}

\appendix

\newpage

\section{Appendices}
\label{sec:appendix}

\subsection{Mean performance}
\label{sec:overal-mean-performance}

% Mean performance result on train and dev/test set
\begin{table}[!h]
\setlength{\tabcolsep}{4.0pt}
\centering
\small{
\begin{tabular}
{ c||c||c|c|c}
\hline
\textbf{Task} & \textbf{Model} & \textbf{Train} & \textbf{Dev.} & \textbf{Test Avg.} \\ \hline \hline
\multirow{3}{4em}{Zh$\Rightarrow$En} & rnn & 35.02 & 41.02 & 37.73 \\ 
& fconv & 40.02 & 45.58 & 43.04 \\ 
& san   & 38.46 & 47.85 & 45.17 \\ \hline \hline
\multirow{3}{4em}{En$\Rightarrow$Zh} & rnn & 39.12 & 22.57 & 16.61 \\ 
& fconv & 40.91 & 24.96 & 18.43 \\ 
& san   & 41.67 & 26.31 & 19.50 \\  \hline
\end{tabular}
}
\caption{The average sense generalization performance results on NIST benchmark measured by BLEU; note that here \textbf{Train} is measured through single reference while \textbf{Dev.} is measured by four references for the Zh$\Rightarrow$En task, so for rnn, \textbf{Dev.} can surpass \textbf{Train}.}
\label{tab:model-performance}
\end{table}

\subsection{Evaluation metrics}
\label{sec:eval-metrics}

\noindent
\textbf{overlap@$k$}
The first metric we use for evaluating the accuracy
of the estimated risk is based on the overlap@$k$
metric~\citep{dong-etal-2018-confidence}. Since
each source word $\tn{x}_i$ is annotated with a risk $r_i$
via exactly or approximately generating couterfactuals.
The risks then induce a ranking among the source words.
According to our Definition~\ref{def:generalization-barrier-words},
the top-$k$ risky words are treated as generalization barrier
words. So given two rankings of the same input, we can choose
their top-$k$ risky words and measure how they overlap with each
other. Formally, given two ranked list of words of the input $\tn{x}$
based on two list of risks,
$\tau_1$ and $\tau_2$ are their top-$k$ risky words,
the overlap@$k$ metric is as follows:
\begin{equation}
    \text{overlap@}k = \frac{\tau_1 \cap \tau_2}{k}
\end{equation}

\noindent
\textbf{Kendall's coefficient concordance}
The second metric for evaluting rank stability (variance) is called Kendall’s coefficient of concordance~\citep{Mazurek2011EVALUATIONOR}. It is computed through the following formula:
\begin{equation}
    W = \frac{
        \sum_{i=1}^n X_i^2 - \frac{(\sum_i^n X_i)^2}{n}
    }{
        \frac{1}{12} \cdot k^2 \cdot (n^3 - n)
    },
\end{equation}
where $k$ is the number of rankings and $n$ the number of objects. In our setting, $k$ is 25 corresponding to the 25 repeats of the simulation and $n$ is the source sentence length corresponding to the length of ranks on all the source words.

\subsection{Definition of troublesome words}
\label{sec:detailed-info-for-trouble-makers}

In Section~\ref{sec:compare-with-other-source-cat},
we measure the similarity between
our identified generalization barrier words and previouly proposed
under-translated words~\citep{zhao2019addressing}
and troublesome words~\citep{zhao2018addressing}.
Here, we give a detailed introduction to the definition of them.

\noindent
\textbf{Under-translated words}
The under-translated word $v \in \mathcal{V}^s$~\citep{zhao2019addressing}
is defined as the word with its translation entropy
larger than certain threshold.
Each word's translation entropy is calculated
from its translation probabilities $\phi(w \vert v)$ which are
count-based estimated from word alignments
of the training set obtained through certain statistical word aligner,
e.g. $\text{fast\_align}$~\citep{dyer-etal-2013-simple}.
That is, for each $v \in \mathcal{V}^s$,
$te(v) =  \sum_{w} - \phi(w \vert v) \cdot \log \phi(w \vert v)$,
where $w \in \mathcal{V}'$.
So we can use $te(v)$ of each word to annotate each source
sentence with every word with a global risk.

\noindent
\textbf{Troublesome words}
The troublesome word $v$~\citep{zhao2018addressing} is defined
as word that satisfies certain exception condition, which
is measured through an exception rate $er(v) = \frac{M^v}{N^v}$.
Here, $N^v$ is the number of alignment pair $(v, w)$
for any $w \in \mathcal{V}'$,
across the whole corpus obtained as well with $\text{fast\_align}$;
$M^v$ is the number of exception alignment pair where $w$ has
violated certain conditions. \citet{zhao2018addressing} proposes
three exception conditions which result in similar performance,
so here we use only one of them for experiment. That is, the
word probability $P_{\hat{\theta}}(\tn{y}_t = w \vert \tn{y}_{<t}, \tn{x})$
falls below certain threshold $p_0$.
The same with the under-translated word, we use $er(v)$ to label
each source word.

\subsection{Measures for evaluating the re-ranking candidates}
\label{sec:metrics-for-reranking}

In Section~\ref{sec:potential-usage}, we use three measures
to characterize the candidates generated by top-1 beam search from
several randomly edited sources
via barrier words and commonly used top-$k$ beam search results
from the original source input. Here, we give a detailed
description of those measures. We denote the hypo candidates
generated from source-editing top-1 beam search and
top-$k$ beam search as $\mathcal{C}_1$ and $\mathcal{C}_2$.
To be fair, the two collections of hypo candidates have same size, that is
$\vert \mathcal{C}_ \vert = \vert \mathcal{C}_2 \vert$.

\noindent
\textbf{Oracle} Given the reference $\tn{y}^*$,
a set of candidates $\mathcal{C}_i$ ($i \in \{1, 2\}$),
the oracle value of $\mathcal{C}_i$ is:
\begin{equation}
    \mathcal{O}(\mathcal{C}_i, \tn{y}^*) = \max_{\hat{\tn{y}} \in \mathcal{C}_i} \text{BLEU}(\hat{\tn{y}}, \tn{y}^*),
\end{equation}
where the function $\text{BLEU}$ denotes the sentence-level
smoothed BLEU~\citep{Lin2004AutomaticEO} in all our experiments.
The larger the oracle value is, the better the candidates are.

\noindent
\textbf{Coverage} Given the reference $\tn{y}^*$,
a set of candidates $\mathcal{C}_i$ ($i \in \{1, 2\}$),
the coverage value of $\mathcal{C}_i$ is:
\begin{equation}
    \mathcal{C}(\mathcal{C}_i, \tn{y}^*) = \frac{\text{1-Gram}(\tn{y}^*) \cap \cup_{\hat{\tn{y}} \in \mathcal{C}_i}\text{1-Gram}(\mathcal{\hat{\tn{y}}})}{\text{1-Gram}(\tn{y}^*)},
\end{equation}
where $\text{1-Gram}(\cdot)$ denotes the different 1-grams of the sentence $\cdot$.
The layer the coverage value is, the better the candidates are.

\noindent
\textbf{Diversity} Given
a set of candidates $\mathcal{C}_i$ ($i \in \{1, 2\}$),
the diversity value of $\mathcal{C}_i$ is:
\begin{equation}
    \mathcal{D}(\mathcal{C}_i) = \frac{1}{\vert \mathcal{C}_i \vert * (\vert \mathcal{C}_i - 1 \vert)}
    \sum_{\hat{\tn{y}} \in \mathcal{C}_i, \hat{\tn{y}}' \in \mathcal{C}_i} \text{BLEU}(\hat{\tn{y}}, \hat{\tn{y}}'),
\end{equation}
where $\hat{\tn{y}} \neq \hat{\tn{y}}'$. That is we use the sentence-level smoothed BLEU for comparing
the difference between any two candidates and average them all. So the smaller the diversity value is,
the better the candidates are.

\end{document}